\documentclass[10pt, a4paper]{article}
\usepackage{lrec}
\usepackage{graphicx}
\usepackage{tabularx}
\usepackage{soul}
\usepackage{epstopdf}
\usepackage[latin1]{inputenc}
\usepackage{hyperref}

\usepackage{xstring}
\usepackage{booktabs}
\usepackage{ragged2e}
\usepackage{lipsum}
\usepackage{verbatim}
\usepackage{multirow}
\usepackage{footmisc}
\usepackage{caption}
\usepackage{subcaption}
\usepackage{siunitx}
\usepackage{adjustbox}
\usepackage{footnote}
\makesavenoteenv{tabular}
\makesavenoteenv{table}

\usepackage{tikz}
\usetikzlibrary{shapes.geometric, arrows}
\tikzstyle{box} = [rectangle, rounded corners, minimum width=0.75cm, minimum height=0.75cm,text centered, draw=black, fill=blue!10]

\tikzstyle{box4text} = [rectangle, rounded corners, minimum width=0.75cm, minimum height=0.75cm,text centered, draw=white, fill=white]

\tikzstyle{arrow} = [thick,->,>=stealth]

\usepackage{amsmath}
\usepackage{amsfonts}
\usepackage{amssymb}
\usepackage{xspace}
\newcommand{\frbert}{FlauBERT\xspace}
\newcommand{\frbertbase}{$\text{FlauBERT}_\textsc{base}$\xspace}
\newcommand{\frbertlarge}{$\text{FlauBERT}_\textsc{large}$\xspace}

\newcommand{\bert}{BERT\xspace}
\newcommand{\bertbase}{$\text{BERT}_\textsc{base}$\xspace}
\newcommand{\bertlarge}{$\text{BERT}_\textsc{large}$\xspace}

\newcommand{\roberta}{RoBERTa\xspace}
\newcommand{\robertbase}{$\text{RoBERTa}_\textsc{base}$\xspace}
\newcommand{\robertlarge}{$\text{RoBERTa}_\textsc{large}$\xspace}

\newcommand{\cmbert}{CamemBERT\xspace}

\newcommand{\xlmrbase}{$\text{XLM-R}_\textsc{base}$\xspace}
\newcommand{\xlmrlarge}{$\text{XLM-R}_\textsc{large}$\xspace}

\makeatletter
\DeclareRobustCommand\onedot{\futurelet\@let@token\@onedot}
\def\@onedot{\ifx\@let@token.\else.\null\fi\xspace}
\def\eg{\emph{e.g}\onedot} 
\def\ie{\emph{i.e}\onedot} 
 
\def\etc{\emph{etc}\onedot} 
\def\wrt{\emph{w.r.t}\onedot} 

\makeatother

\usepackage{xcolor}

\definecolor{teamwsdcolor}{rgb}{0.0, 0.7, 0.5}

\title{\frbert: Unsupervised Language Model Pre-training for French}

\name{\normalsize\bfseries Hang Le\textsuperscript{1}\qquad Lo\"{i}c Vial\textsuperscript{1}\qquad Jibril Frej\textsuperscript{1}\qquad Vincent Segonne\textsuperscript{2}\qquad Maximin Coavoux\textsuperscript{1}\\\bfseries Benjamin Lecouteux\textsuperscript{1}\qquad Alexandre Allauzen\textsuperscript{3}\qquad Beno\^it Crabb\'e\textsuperscript{2}\qquad Laurent Besacier\textsuperscript{1} \qquad Didier Schwab\textsuperscript{1} \\}

\address{\small\textsuperscript{1}Univ. Grenoble Alpes, CNRS, LIG\qquad \textsuperscript{2}Universit\'e Paris Diderot\qquad\textsuperscript{3}E.S.P.C.I, CNRS LAMSADE, PSL Research University  \\ 
	\footnotesize\ttfamily
	\{thi-phuong-hang.le, loic.vial, jibril.frej\}@univ-grenoble-alpes.fr\\
	\footnotesize\ttfamily
	\{maximin.coavoux, didier.schwab, benjamin.lecouteux, laurent.besacier\}@univ-grenoble-alpes.fr\\
	\footnotesize\ttfamily
	\{vincent.segonne@etu, bcrabbe@linguist\}.univ-paris-diderot.fr, alexandre.allauzen@espci.fr
}
\abstract{Language models have become a key step to achieve state-of-the art results in many different Natural Language Processing (NLP) tasks. Leveraging the huge amount of unlabeled texts nowadays available, they provide an efficient way to pre-train continuous word representations that can be fine-tuned for a downstream task, along with their contextualization at the sentence level.
This has been widely demonstrated for English using contextualized representations \cite{dai2015semi,peters2018deep,howard2018universal,radford2018improving,devlin2019bert,yang2019xlnet}.
In this paper, we introduce and share \frbert, a model learned on a very large and heterogeneous French corpus. Models of different sizes are trained using the new CNRS (French National Centre for Scientific Research) \textit{Jean Zay} supercomputer. We apply our French language models to diverse NLP tasks (text classification, paraphrasing, natural language inference, parsing, word sense disambiguation) and show that most of the time they outperform other pre-training approaches. Different versions of \frbert as well as a unified evaluation protocol for the downstream tasks, called FLUE (French Language Understanding Evaluation), are shared to the research community for further reproducible experiments in French NLP.\\
\newline  }

\begin{document}

\maketitleabstract



\section{Introduction}
A recent game-changing contribution in Natural Language Processing (NLP) was the introduction of deep \emph{unsupervised} language representations pre-trained using only plain text corpora. Previous word embedding pre-training approaches, such as word2vec \cite{Mikolov:2013:DRW:2999792.2999959} or GloVe \cite{Pennington14glove:global}, learn a single vector for each wordform.
By contrast, these new models are trained to produce \textit{contextual embeddings}: the output representation depends on the entire input sequence (\eg each token instance has a vector representation that depends on its left and right context). Initially based on recurrent neural networks \cite{dai2015semi,ramachandran2017unsupervised,howard2018universal,peters2018deep}, these models quickly converged towards the use of the Transformer \cite{vaswani2017attention}, such as GPT \cite{radford2018improving}, BERT \cite{devlin2019bert}, XLNet \cite{yang2019xlnet}, RoBERTa \cite{liu2019roberta}. Using these pre-trained models in a transfer learning fashion has shown to yield striking improvements across a wide range of NLP tasks. One can easily build state-of-the-art NLP systems thanks to the publicly available pre-trained weights, saving time, energy, and resources.
As a consequence, unsupervised language model pre-training has become a \textit{de facto} standard in NLP. This has been, however, mostly demonstrated for English even though multi-lingual or cross-lingual variants are also available, taking into account more than a hundred languages in a single model: mBERT \cite{devlin2019bert}, XLM \cite{lample2019cross}, XLM-R \cite{conneau2019unsupervised}.


In this paper, we describe our methodology to build \frbert{} -- \textbf{F}rench \textbf{La}nguage \textbf{U}nderstanding via \textbf{B}idirectional \textbf{E}ncoder \textbf{R}epresentations from \textbf{T}ransformers, a French BERT\footnote{We learned of a similar project that resulted in a publication on arXiv \cite{martin2019camembert}. However, we believe that these two works on French language models are complementary since the NLP tasks we addressed are different, as are the training corpora and preprocessing pipelines. We also point out that our models were trained using the CNRS (French National Centre for Scientific Research) public research computational infrastructure and did not receive any assistance from a private stakeholder.} model that outperforms multi-lingual/cross-lingual models in several downstream NLP tasks, under similar configurations. \frbert relies on freely available datasets and is made publicly available in different versions.\footnote{\url{https://github.com/getalp/Flaubert}} For further reproducible experiments, we also provide the complete processing and training pipeline as well as a general benchmark for evaluating French NLP systems. This evaluation setup is similar to the popular GLUE benchmark \cite{wang-etal-2018-glue}, and is named FLUE (French Language Understanding Evaluation).

\section{Related Work}

\subsection{Pre-trained Language Models}
Self-supervised\footnote{\emph{Self-supervised learning} is a special case of \emph{unsupervised learning} where unlabeled data is used as a supervision signal.} pre-training on unlabeled text data was first proposed in the task of neural language modeling \cite{bengio2003neural,collobert2008unified}, where it was shown that a neural network trained to predict next word from prior words can learn useful embedding representations, called \emph{word embeddings} (each word is represented by a fixed vector). These representations were shown to play an important role in NLP, yielding state-of-the-art performance on multiple tasks \cite{collobert2011natural}, especially after the introduction of word2vec \cite{Mikolov:2013:DRW:2999792.2999959} and GloVe \cite{Pennington14glove:global}, efficient and effective algorithms for learning word embeddings.

A major limitation of word embeddings is that a word can only have a single representation, even if it can have multiple meanings (\eg depending on the context). Therefore, recent works have introduced a paradigm shift from context-free word embeddings to \emph{contextual embeddings}: the output representation is a function of the entire input sequence, which allows encoding complex, high-level syntactic and semantic characteristics of words or sentences.

This line of research was started by~\newcite{dai2015semi} who proposed pre-training representations via either an encoder-decoder language model or a sequence autoencoder. \newcite{ramachandran2017unsupervised}\footnote{It should be noted that learning contextual embeddings was also proposed in \cite{mccann2017learned}, but in a \emph{supervised} fashion as they used annotated machine translation data.} showed that this approach can be applied to pre-training sequence-to-sequence models \cite{sutskever2014sequence}. These models, however, require a significant amount of in-domain data for the pre-training tasks. \newcite[ELMo]{peters2018deep} and \newcite[ULMFiT]{howard2018universal} were the first to demonstrate that leveraging huge general-domain text corpora in pre-training can lead to substantial improvements on downstream tasks. Both methods employ LSTM \cite{hochreiter1997long} language models, but ULMFiT utilizes a regular multi-layer architecture, while ELMo adopts a \emph{bidirectional} LSTM to build the final embedding for each input token from the concatenation of the left-to-right and right-to-left representations. Another fundamental difference lies in how each model can be tuned to different downstream tasks: ELMo delivers different word vectors that can be interpolated, whereas ULMFiT enables robust fine-tuning of the whole network \wrt the downstream tasks. The ability of fine-tuning was shown to significantly boost the performance, and thus this approach has been further developed in the recent works such as MultiFiT \cite{eisenschlos2019multifit} or most prominently Transformer-based \cite{vaswani2017attention} architectures: GPT \cite{radford2018improving}, BERT \cite{devlin2019bert}, XLNet \cite{yang2019xlnet}, XLM \cite{lample2019cross}, RoBERTa \cite{liu2019roberta}, ALBERT \cite{lan2019albert}, T5 \cite{raffel2019exploring}. These methods have one after the other established new state-of-the-art results on various NLP benchmarks, such as GLUE \cite{wang-etal-2018-glue} or SQuAD \cite{rajpurkar2018know}, surpassing previous methods by a large margin.

\subsection{Pre-trained Language Models Beyond English}


Given the impact of pre-trained language models on NLP downstream tasks in English, several works have recently released pre-trained models for other languages. 
For instance, ELMo exists for Portuguese, Japanese, German and Basque,\footnote{\url{https://allennlp.org/elmo}} while BERT and variants were specifically trained for simplified and traditional Chinese\footref{footnote:google-bert}
and German.\footnote{\url{https://deepset.ai/german-bert}} A Portuguese version of MultiFiT is also available.\footnote{\label{footnote:multifit}\url{https://github.com/piegu/language-models}} Recently, more monolingual BERT-based models have been released, such as for Arabic \cite{antoun2020arabert}, Dutch \cite{de2019bertje,delobelle2020robbert}, Finnish \cite{virtanen2019multilingual}, Italian \cite{PolignanoEtAlCLIC2019}, Portuguese \cite{souza2019portuguese}, Russian \cite{kuratov2019adaptation}, Spanish \cite{CaneteCFP2020}, and Vietnamese \cite{nguyen2020phobert}. For French, besides pre-trained language models using ULMFiT and MultiFiT configurations,\footref{footnote:multifit} \cmbert \cite{martin2019camembert} is a French BERT model concurrent to our work.

Another trend considers one model estimated for several languages with a shared vocabulary.
The release of multilingual \bert for 104 languages  pioneered this approach.\footnote{\label{footnote:google-bert}\url{https://github.com/google-research/bert}}
A recent extension of this work leverages parallel data to build a cross-lingual pre-trained version of LASER \cite{artetxe2019massively} for 93 languages, XLM \cite{lample2019cross} and XLM-R \cite{conneau2019unsupervised} for 100 languages. 


\subsection{Evaluation Protocol for French NLP Tasks} 
The existence of a multi-task evaluation benchmark such as GLUE \cite{wang-etal-2018-glue} for English is highly beneficial to facilitate research in the language of interest. The GLUE benchmark has become a prominent framework to evaluate the performance of NLP models in English. The recent contributions based on  pre-trained language models have led to remarkable performance across a wide range of Natural Language Understanding (NLU) tasks. The authors of GLUE have therefore  introduced SuperGLUE \cite{wang2019superglue}:  a new benchmark built on the principles of GLUE, including more challenging and diverse set of tasks. A Chinese version of GLUE\footnote{\url{https://github.com/chineseGLUE/chineseGLUE}} is also developed to evaluate model performance in Chinese NLP tasks. As of now, we have not learned of any such benchmark for French. 

\section{Building \frbert}
In this section, we describe the training corpus, the text preprocessing pipeline, the model architecture and training configurations to build \frbertbase and \frbertlarge. 

\begin{table*}[htb!]
	\centering
	\resizebox{\textwidth}{!}{
		\centering
		\begin{tabular}{l|c|c|c|c}
			\toprule
			& \bertbase & \robertbase & \cmbert & \frbertbase/\frbertlarge \\
			\midrule
			Language & English & English & French & French \\
			Training data & 13 GB & 160 GB & 138 GB\textsuperscript{$\dagger$} & 71 GB\textsuperscript{$\ddagger$} \\
			Pre-training objectives & NSP and MLM & MLM & MLM & MLM \\
			Total parameters & 110 M& 125 M & 110 M & 138 M/ 373 M\\
			Tokenizer & WordPiece 30K & BPE 50K & SentencePiece 32K & BPE 50K\\
			Masking strategy & Static + Sub-word masking & Dynamic + Sub-word masking & Dynamic + Whole-word masking & Dynamic + Sub-word masking\\
			\bottomrule
		
		\multicolumn{5}{l}{\textsuperscript{$\dagger$}, \textsuperscript{$\ddagger$}: 282 GB, 270 GB before filtering/cleaning.}
		\end{tabular}
	}
	\caption{\label{tbl:model_comparison} Comparison between \frbert and previous work.}
	
\end{table*}

\subsection{Training Data}
\paragraph{Data collection} Our French text corpus consists of 24 sub-corpora gathered from different sources, covering diverse topics and writing styles, ranging from formal and well-written text (\eg Wikipedia and books)\footnote{\url{http://www.gutenberg.org}} to random text crawled from the Internet (\eg Common Crawl).\footnote{\url{http://data.statmt.org/ngrams/deduped2017}} The data were collected from three main sources: (1) monolingual data for French provided in WMT19 shared tasks \cite[4 sub-corpora]{li2019findings}; (2) French text corpora offered in the OPUS collection \cite[8 sub-corpora]{TIEDEMANN12.463}; and (3) datasets available in the Wikimedia projects \cite[8 sub-corpora]{wiki:xxx}.

We used the WikiExtractor tool\footnote{\url{https://github.com/attardi/wikiextractor}} to extract the text from Wikipedia. For the other sub-corpora, we either used our own tool to extract the text or download them directly from their websites. The total size of the uncompressed text before preprocessing is 270 GB. More details can be found in Appendix~\ref{appx:desc_datasets}.

\paragraph{Data preprocessing} 
For all sub-corpora, we filtered out very short sentences as well as repetitive and non-meaningful content such as telephone/fax numbers, email addresses, \etc For Common Crawl, which is our largest sub-corpus with 215 GB of raw text, we applied aggressive cleaning to reduce its size to 43.4 GB. All the data were Unicode-normalized in a consistent way before being tokenized using Moses tokenizer \cite{koehn2007moses}. The resulting training corpus is 71 GB in size.

Our code for downloading and preprocessing data is made publicly available.\footnote{\url{https://github.com/getalp/Flaubert}}

		


\subsection{Models and Training Configurations}\label{sec:model-training-configuration}

\paragraph{Model architecture}
\frbert has the same model architecture as \bert \cite{devlin2019bert}, which consists of a multi-layer bidirectional Transformer \cite{vaswani2017attention}. Following \newcite{devlin2019bert}, we propose two model sizes:
\begin{itemize}
	\item \frbertbase: $L=12, H=768, A=12$,
    \item \frbertlarge: $L=24, H=1024, A=16$,
\end{itemize}
where $L, H$ and $A$ respectively denote the number of Transformer blocks, the hidden size, and the number of self-attention heads. As Transformer has become quite standard, we refer to \newcite{vaswani2017attention} for further details.

\paragraph{Training objective and optimization}

Pre-training of the original BERT \cite{devlin2019bert} consists of two supervised tasks: (1) a \emph{masked language model} (MLM) that learns to predict randomly masked tokens; and (2) a \emph{next sentence prediction} (NSP) task in which the model learns to predict whether \texttt{B} is the actual next sentence that follows \texttt{A}, given a pair of input sentences \texttt{A,B}.

\newcite{devlin2019bert} observed that removing NSP significantly hurts performance on some downstream tasks. However, the opposite  was shown in later studies, including \newcite[XLNet]{yang2019xlnet}, \newcite[XLM]{lample2019cross}, and \newcite[RoBERTa]{liu2019roberta}.\footnote{\newcite{liu2019roberta} hypothesized that the original BERT implementation may only have removed the loss term while still retaining a bad input format, resulting in performance degradation.} Therefore, we only employed the MLM objective in \frbert.

To optimize this objective function, we followed \newcite{liu2019roberta} and used the Adam optimizer \cite{kingma2014adam} with the following parameters: 
\begin{itemize}
    \item \frbertbase: warmup steps of 24k, peak learning rate of \num{6e-4}, $\beta_1 = 0.9$, $\beta_2 = 0.98$, $\epsilon = \num{1e-6}$ and weight decay of 0.01.
    \item \frbertlarge: warmup steps of 30k, peak learning rate of \num{3e-4}, $\beta_1 = 0.9$, $\beta_2 = 0.98$, $\epsilon = \num{1e-6}$ and weight decay of 0.01.
\end{itemize}

\paragraph{Training \frbertlarge}

Training very deep Transformers is known to be susceptible to instability \cite{wang2019learning,nguyen2019transformers,xu2019deep,fan2019reducing}. Not surprisingly, we also observed this difficulty when training \frbertlarge using the same configurations as \bertlarge and \robertlarge, where divergence happened at an early stage.

Several methods have been proposed to tackle this issue. For example, in an updated implementation of the Transformer~\cite{vaswani2018tensor2tensor}, layer normalization is applied \emph{before} each attention layer by default, rather than after each residual block as in the original implementation~\cite{vaswani2017attention}. These configurations are called \emph{pre-norm} and \emph{post-norm}, respectively. It was observed by~\newcite{vaswani2018tensor2tensor}, and again confirmed by later works \eg{} \cite{wang2019learning,xu2019deep,nguyen2019transformers}, that pre-norm helps stabilize training. 
Recently, a regularization technique called \emph{stochastic depths}~\cite{huang2016deep} has been demonstrated to be very effective for training deep Transformers, by \eg \newcite{pham2019very} and \newcite{fan2019reducing} who successfully trained architectures of more than 40 layers. The idea is to randomly drop a number of (attention) layers at each training step. Other techniques are also available such as progressive training \cite{gong2019efficient}, or improving initialization \cite{zhang2019fixup,xu2019deep} and normalization \cite{nguyen2019transformers}.

For training \frbertlarge, we employed pre-norm attention and stochastic depths for their simplicity. We found that these two techniques were sufficient for successful training. We set the rate of layer dropping to \num{0.2} in all the experiments.

    
    
    

    
    
    
    
    

    

\paragraph{Other training details} 

A vocabulary of 50K sub-word units is built using the Byte Pair Encoding (BPE) algorithm \cite{sennrich2016neural}. The only difference between our work and \roberta is that the training data are preprocessed and tokenized using a basic tokenizer for French \cite[Moses]{koehn2007moses}, as in XLM \cite{lample2019cross}, before the application of BPE. We use \textit{fastBPE},\footnote{\url{https://github.com/glample/fastBPE}} a very efficient implementation to extract the BPE units and encode the corpora. 

\frbertbase is trained on 32 GPUs Nvidia V100 in 410 hours and \frbertlarge is trained on 128 GPUs in 390 hours, both with the effective batch size of 8192 sequences. 

Finally, we summarize the differences between \frbert and \bert, \roberta, \cmbert in Table~\ref{tbl:model_comparison}.



\section{FLUE}
In this section, we compile a set of existing French language tasks to form an evaluation benchmark for French NLP that we called FLUE (French Language Understanding Evaluation). 
We select the datasets from different domains, level of difficulty, degree of formality, and amount of training samples. Three out of six tasks (Text Classification, Paraphrase, Natural Language Inference) are from cross-lingual datasets since we also aim to provide results from a monolingual pre-trained model to facilitate future studies of cross-lingual models, which have been drawing much of research interest recently.

Table~\ref{tbl:tasks_data} gives an overview of the datasets, including their domains and training/development/test splits. The details are presented in the next subsections.

\begin{table}[htb!]
	\centering
	\resizebox{\columnwidth}{!}{
		\centering
		\begin{tabular}{lrlccc}
			\toprule
			\multicolumn{2}{c}{Dataset} & Domain & Train & Dev & Test \\
			\midrule
			\multirow{3}{*}{CLS-$\texttt{FR}$} & Books &  \multirow{3}{*}{Product reviews} & 2\,000 & - & 2\,000\\
			& DVD & & 1\,999 & - & 2\,000 \\
			& Music & & 1\,998 & - & 2\,000\\
			\midrule
            \multicolumn{2}{l}{PAWS-X-$\texttt{FR}$}  & General domain
            & 49\,401 & 1\,992 & 1\,985 \\
            \midrule
            \multicolumn{2}{l}{XNLI-$\texttt{FR}$}  & Diverse genres
            & 392\,702 & 2\,490 & 5\,010\\
            \midrule
            \multicolumn{2}{l}{French Treebank}
            & Daily newspaper  
            & 14\,759 & 1\,235 & 2\,541 \\
            \midrule
            \multicolumn{2}{l}{FrenchSemEval}   & Diverse genres & 55\,206 &- &3\,199 \\
            \midrule
            \multicolumn{2}{l}{Noun Sense Disambiguation} & Diverse genres
            & 818\,262 & - & 1\,445 \\
			\bottomrule
		
        
		\end{tabular}
	}
	\caption[caption]{\label{tbl:tasks_data} Descriptions of the datasets included in our FLUE benchmark.
	}

\end{table}

\subsection{Text Classification}
\paragraph{CLS} The Cross Lingual Sentiment  CLS \cite{prettenhofer2010cross} dataset consists of Amazon reviews for three product categories: books, DVD, and music in four languages: English, French, German, and Japanese. Each sample contains a review text and the associated rating from 1 to 5 stars. Following \newcite{blitzer2006domain} and \newcite{prettenhofer2010cross}, ratings with 3 stars are removed. Positive reviews have ratings higher than 3 and negative reviews are those rated lower than 3. There is one train and test set for each product category. The train and test sets are balanced, including around 1\,000 positive and 1\,000 negative reviews for a total of 2\,000 reviews in each dataset. We take the French portion to create the binary text classification task in FLUE and report the accuracy on the test set.

\subsection{Paraphrasing}
\paragraph{PAWS-X} The Cross-lingual Adversarial Dataset for Paraphrase Identification PAWS-X \cite{yang2019pawsx} is the extension of the Paraphrase Adversaries from Word Scrambling PAWS \cite{zhang2019paws} for English to six other languages: French, Spanish, German, Chinese, Japanese and Korean. PAWS composes English paraphrase identification pairs from Wikipedia and Quora in which two sentences in a pair have high lexical overlap ratio, generated by LM-based word scrambling and back translation followed by human judgement. The paraphrasing task is to identify whether the sentences in these pairs are semantically equivalent or not. Similar to previous approaches to create multilingual corpora, \newcite{yang2019pawsx} used machine translation to create the training set for each target language in PAWS-X from the English training set in PAWS. The development and test sets for each language are translated by human translators. We take the related datasets for French to perform the paraphrasing task and report the accuracy on the test set.

\subsection{Natural Language Inference}
\paragraph{XNLI} The Cross-lingual NLI  (XNLI) corpus \cite{conneau2018xnli} extends the development and test sets of the Multi-Genre Natural Language Inference corpus \cite[MultiNLI]{williams2018broad} to 15 languages. The development and test sets for each language consist of 7\,500 human-annotated examples, making up a total of 112\,500 sentence pairs annotated with the labels \textit{entailment}, \textit{contradiction}, or \textit{neutral}. Each sentence pair includes a premise ($p$) and a hypothesis ($h$). The Natural Language Inference (NLI) task, also known as recognizing textual entailment (RTE), is to determine whether $p$ entails, contradicts or neither entails nor contradicts $h$. We take the French part of the XNLI corpus to form the development and test sets for the NLI task in FLUE. The train set is obtained from the machine translated version to French provided in XNLI. Following \newcite{conneau2018xnli}, we report the test accuracy.

\subsection{Parsing and Part-of-Speech Tagging}

Syntactic parsing consists in assigning a tree structure to a sentence in natural language.
We perform parsing on the French Treebank \cite{Abeille2003}, a collection of sentences extracted from French daily newspaper Le Monde, and manually annotated with both constituency and dependency syntactic trees and part-of-speech tags.
Specifically, we use the version of the corpus instantiated for the SPMRL 2013 shared task and described by \newcite{seddah-etal-2013-overview}.
This version is provided with a standard split representing 14\,759 sentences for the training corpus, and respectively 1\,235 and 2\,541 sentences for the development and evaluation sets.


\subsection{Word Sense Disambiguation Tasks}
Word Sense Disambiguation (WSD) is a classification task which aims to predict the sense of words in a given context according to a specific sense inventory. We used two French WSD tasks: the FrenchSemEval task \cite{segonne2019using}, which targets verbs only, and a modified version of the French part of the Multilingual WSD task of SemEval 2013 \cite{Navigli2013}, which targets nouns.

\paragraph{Verb Sense Disambiguation}
We made experiments of sense disambiguation focused on French verbs using FrenchSemEval \cite[FSE]{segonne2019using}, an evaluation dataset in which verb occurrences were manually sense annotated with the sense inventory of Wiktionary, a collaboratively edited open-source dictionary. FSE includes both the evaluation data and the sense inventory. The evaluation data consists of 3\,199 manual annotations among a selection of 66 verbs which makes roughly 50 sense annotated occurrences per verb. The sense inventory provided in FSE is a Wiktionary dump (04-20-2018) openly available via Dbnary \cite{serasset2012dbnary}. For a given sense of a target key, the sense inventory offers a definition along with one or more examples. For this task, we considered the examples of the sense inventory as training examples and tested our model on the evaluation dataset.

\paragraph{Noun Sense Disambiguation}


We propose a new challenging task for the WSD of French, based on the French part of the Multilingual WSD task of SemEval 2013 \cite{Navigli2013}, which targets nouns only. We adapted the task to use the WordNet 3.0 sense inventory \cite{miller1995wordnet} instead of BabelNet \cite{navigli2010babelnet}, by converting the sense keys to WordNet 3.0 if a mapping exists in BabelNet, and removing them otherwise.

The result of the conversion process is an evaluation corpus composed of 306 sentences and 1\,445 French nouns annotated with WordNet sense keys, and manually verified. 

For the training data, we followed the method proposed by \newcite{hadjsalahthese2018}, and translated the SemCor \cite{Miller1993} and the WordNet Gloss Corpus\footnote{The set of WordNet glosses semi-automatically sense annotated which is released as part of WordNet since version 3.0.} into French, using the best English-French Machine Translation system of the \emph{fairseq} toolkit\footnote{\url{https://github.com/pytorch/fairseq}} \cite{ott2019fairseq}. Finally, we aligned the WordNet sense annotation from the source English words to the the translated French words, using the alignment provided by the MT system.

We rely on WordNet sense keys instead of the original BabelNet annotations for the following two reasons.
First, WordNet is a resource that is entirely manually verified, and widely used in WSD research \cite{Navigli2009WSD}.
Second, there is already a large quantity of sense annotated data based on the sense inventory of WordNet \cite{vialhal01718237} that we can use for the training of our system.

We publicly release\footnote{\url{https://zenodo.org/record/3549806}} both our training data and the evaluation data in the UFSAC format \cite{vialhal01718237}.

\section{Experiments and Results}
In this section, we present \frbert fine-tuning results on the FLUE benchmark. We compare the performance of \frbert with Multilingual BERT \cite[mBERT]{devlin2019bert} and \cmbert \cite{martin2019camembert} on all tasks. In addition, for each task we also include the best non-BERT model for comparison. We made use of the open source libraries \cite[XLM]{lample2019cross} and \cite[Transformers]{Wolf2019HuggingFacesTS} in some of the experiments.

\subsection{Text Classification}
\label{sec:finetune_clf}
\paragraph{Model description} We followed the standard fine-tuning process of \bert \cite{devlin2019bert}. The input is a degenerate text-$\varnothing$ pair. The classification head is composed of the following layers, in order: dropout, linear, $\tanh$ activation, dropout, and linear. The output dimensions of the linear layers are respectively equal to the hidden size of the Transformer and the number of classes (which is $2$ in this case as the task is binary classification). The dropout rate was set to $0.1$.

We trained for 30 epochs using a batch size of 16 while performing a grid search over 4 different learning rates: \num{1e-5}, \num{5e-5}, \num{1e-6}, and \num{5e-6}. A random split of 20\% of the training data was used as validation set, and the best performing model on this set was then chosen for evaluation on the test set.

\begin{table}[!htb]
	\centering
		\centering
		\begin{tabular}{lccc}
			\toprule
			Model & Books & DVD & Music\\
			\midrule
            MultiFiT\textsuperscript{$\dagger$} & 91.25 & 89.55 & 93.40 \\
			mBERT\textsuperscript{$\dagger$} & 86.15 & 86.90 & 86.65 \\
			\cmbert & 92.30 & 93.00 & 94.85 \\
			\frbertbase & 93.10 & 92.45 & 94.10 \\
			\frbertlarge & \textbf{95.00} & \textbf{94.10} & \textbf{95.85} \\
			\bottomrule
    		\multicolumn{4}{l}{\small\textsuperscript{$\dagger$} Results reported in \cite{eisenschlos2019multifit}.}
		
		\end{tabular}
		\caption{\label{tbl:exp_cls} Accuracy on the CLS dataset for French.}
\end{table}

\paragraph{Results} Table~\ref{tbl:exp_cls} presents the final accuracy on the test set for each model. The results highlight the importance of a monolingual French model for text classification: both \cmbert and \frbert outperform mBERT by a large margin. \frbertbase performs moderately better than \cmbert in the books dataset, while its results on the two remaining datasets of DVD and music are lower than those of \cmbert. \frbertlarge achieves the best results in all categories. 


\subsection{Paraphrasing}
\paragraph{Model description} The setup for this task is almost identical to the previous one, except that: (1) the input sequence is now a pair of sentences \texttt{A,B}; and (2) the hyper-parameter search is performed on the development data set (\ie no validation split is needed).

\paragraph{Results} The final accuracy for each model is reported in Table~\ref{tbl:exp_pawsx}. One can observe that the monolingual French models perform only slightly better than the multilingual model mBERT, which could be attributed to the characteristics of the PAWS-X dataset. Containing samples with high lexical overlap ratio, this dataset has been proved to be an effective measure of model sensitivity to word order and syntactic structure \cite{yang2019pawsx}. A multilingual model such as mBERT, therefore, could capture these features as well as a monolingual model.



\begin{table}[!htb]
	\centering
		\centering
		\begin{tabular}{lc}
			\toprule
			Model & Accuracy  \\
			\midrule
			ESIM\textsuperscript{$\dagger$} \cite{chen2017enhanced} & 66.20 \\
			mBERT\textsuperscript{$\dagger$} & 89.30\\
			\cmbert & \textbf{90.14}\\
			\frbertbase & 89.49 \\
			\frbertlarge & 89.34 \\
			\bottomrule
			
			\multicolumn{2}{l}{\small\textsuperscript{$\dagger$} Results reported in \cite{yang2019pawsx}.}
    		
		\end{tabular}
	\caption{\label{tbl:exp_pawsx} Results on the French PAWS-X dataset.}
\end{table}

\subsection{Natural Language Inference}
\paragraph{Model description} As this task was also considered in \cite[\cmbert]{martin2019camembert}, for a fair comparison, here we replicate the same experimental setup. Similar to paraphrasing, the model input of this task is also a pair of sentences. The classification head, however, consists of only one dropout layer followed by one linear layer.


\paragraph{Results} We report the final accuracy for each model in Table~\ref{tbl:exp_xnli}. The results confirm the superiority of the French models compared to the multilingual model mBERT on this task. \frbertlarge performs moderately better than \cmbert. Both of them clearly outperform \xlmrbase, while cannot surpass \xlmrlarge. 

\begin{table}[!htb]
	\centering
	\begin{tabular}{lc}
		\toprule
		Model & Accuracy  \\
		\midrule
		\xlmrlarge\textsuperscript{$\dagger$} & \textbf{85.2} \\
		\xlmrbase\textsuperscript{$\dagger$} & 80.1 \\
		mBERT\textsuperscript{$\ddagger$} & 76.9\\
		\cmbert\textsuperscript{$\ddagger$} & 81.2 \\
		\frbertbase & 80.6 \\
		\frbertlarge & 83.4 \\
		\bottomrule
		
		\multicolumn{2}{l}{\small\textsuperscript{$\dagger$} Results reported in \cite{conneau2019unsupervised}.} \\
		\multicolumn{2}{l}{\small\textsuperscript{$\ddagger$} Results reported in \cite{martin2019camembert}.}
		
	\end{tabular}
	\caption{\label{tbl:exp_xnli} Results on the French XNLI dataset.}
\end{table}

\subsection{Constituency Parsing and POS Tagging}

\paragraph{Model description}
We use the parser described by \newcite{kitaev-klein-2018-constituency} and \newcite{kitaev-etal-2019-multilingual}.
It is an openly available\footnote{\url{https://github.com/nikitakit/self-attentive-parser}} chart parser based on a self-attentive encoder.
We compare (i) a model without any pre-trained parameters, (ii) a model that additionally uses and fine-tunes fastText\footnote{\url{https://fasttext.cc/}} pre-trained embeddings, (iii) models based on pre-trained language models: mBERT, \cmbert, and \frbert.
We use the default hyperparameters from \newcite{kitaev-klein-2018-constituency} for the first two settings and the hyperparameters from \newcite{kitaev-etal-2019-multilingual} when using pre-trained language models, except for \frbertlarge. For this last model, we use a different learning rate (0.00001), batch size (8) and  ignore training sentences longer than 100 tokens, due to memory limitation.
We jointly perform part-of-speech (POS) tagging based on the same input as the parser, in a multitask setting.
For each setting we perform training 3 times with different random seeds and select best model according
to development F-score.

For final evaluation, we use the evaluation tool provided by the SPMRL shared task organizers\footnote{\url{http://pauillac.inria.fr/~seddah/evalb_spmrl2013.tar.gz}} and report labelled F-score, the standard metric for constituency parsing evaluation, as well as POS tagging accuracy.

\begin{table}[!htb]
    \resizebox{\columnwidth}{!}{
        \begin{tabular}{lcccc}
            \toprule
            Model &  \multicolumn{2}{c}{Dev} & \multicolumn{2}{c}{Test}\\
                            \cmidrule(lr){2-3} \cmidrule(lr){4-5} 
                  & F$_1$ & POS & F$_1$ & POS \\
            \midrule
            Best published \cite{kitaev-etal-2019-multilingual} & & & 87.42 & \\
            \midrule
            No pre-training                                          & 84.31 & 97.6 & 83.85 & 97.5 \\
            fastText pre-trained embeddings                          & 84.09 & 97.6 & 83.64 & 97.7 \\
            mBERT                                                  & 87.25 & 98.1 & 87.52 & 98.1 \\
            \cmbert \cite{martin2019camembert}                    & 88.53 & 98.1 & 88.39 & \textbf{98.2} \\
            \frbertbase                                             & 88.95 & \textbf{98.2} & \textbf{89.05} & 98.1  \\
            \frbertlarge                                           & \textbf{89.08} & \textbf{98.2} & 88.63 & \textbf{98.2} \\
            \midrule
            Ensemble: \frbertbase+ \cmbert                     & \textbf{89.32} & & \textbf{89.28} & \\
            \bottomrule
        \end{tabular}
    }
    \caption{Constituency parsing and POS tagging results.}
    \label{tab:c-parsing}
\end{table}

\paragraph{Results}
We report constituency parsing results in Table~\ref{tab:c-parsing}.
Without pre-training, we replicate the result from \newcite{kitaev-klein-2018-constituency}.
FastText pre-trained embeddings do not bring improvement over this already strong model.
When using pre-trained language models, we observe that \cmbert, with its language-specific
training improves over mBERT by~0.9 absolute F$_1$.
\frbertbase outperforms \cmbert by~0.7 absolute F$_1$ on the test set and obtains
the best published results on the task for a single model.
Regarding POS tagging, all large-scale pre-trained language models obtain similar results (98.1-98.2),
and outperform models without pre-training or with fastText embeddings (97.5-97.7).
\frbertlarge provides a marginal improvement on the development set, and fails to reach \frbertbase results on the test set.

In order to assess whether \frbert and \cmbert are complementary for this task,
we evaluate an ensemble of both models (last line in Table~\ref{tab:c-parsing}).
The ensemble model improves by~0.4 absolute F$_1$ over \frbert on the development set and 0.2 on the test set,
obtaining the highest result for the task.
This result suggests that both pre-trained language models are complementary and have their own strengths and weaknesses.

\subsection{Dependency parsing}
\paragraph{Model} We use our own reimplementation of the parsing model of \newcite{dozat-2017} with maximum spanning tree decoding adapted to handle several input sources such as \bert  representations. The model does not perform part of speech tagging but uses the predicted tags provided by the SPMRL shared task organizers. 

Our word representations are a concatenation of 
word embeddings and tag embeddings learned together with the model parameters on the French Treebank data itself, and at most one of (fastText, \cmbert, \frbertbase, \frbertbase, mBERT) word vector.
As \newcite{dozat-2017}, we use word and tag dropout ($d=0.5$) on word and tag embeddings but without dropout on \bert representations. We performed a fairly comprehensive grid search on hyperparameters for each model tested. 

\begin{table}[htbp]
\begin{center}
\begin{tabular}{lcc}\toprule
Model                 & UAS   & LAS\\\midrule
Best published \cite{constant-2013}&89.19&85.86\\\midrule	
No pre-training       & 88.92 & 85.11\\
fastText pre-training & 86.32 & 82.04\\
mBERT                 &89.50   & 85.86\\
\cmbert             & 91.37 & 88.13\\
\frbertbase& 91.56& 88.35\\
\frbertlarge& {\bf 91.61}& {\bf 88.47}
\\\bottomrule
\end{tabular}
\end{center}
\caption{\label{tab-depparse}Dependency parsing results.}
\end{table}

\paragraph{Results} The results are reported in Table~\ref{tab-depparse}. The best published results in this shared task \cite{constant-2013} were involving an ensemble of parsers with additional resources for modelling multi word expressions (MWE), typical of the French treebank annotations. The monolingual French BERT models (\cmbert, \frbert) perform better and set the new state of the art on this dataset with a single parser and without specific modelling for MWEs. One can observe that both \frbert models perform marginally better than \cmbert, while all of them outperform mBERT by a large margin.

\subsection{Word Sense Disambiguation}

\paragraph{Verb Sense Disambiguation}

Disambiguation was performed with the same WSD supervised method used by \newcite{segonne2019using}. 
First we compute sense vector representations from examples found in the Wiktionary sense inventory: given a sense $s$ and its corresponding examples, we compute the vector representation of $s$ by averaging the vector representations of its examples. 
Then, we tag each test instance with the sense whose representation is the closest based on cosine similarity.
We used the contextual embeddings output by \frbert as vector representations for any given instance (from the sense inventory or the test data) of a target word. We proceeded the same way with mBERT and \cmbert for comparison. We also compared our model with a simpler context vector representation called averaged word embeddings (AWE) which consists in representing  context of target word by averaging its surrounding words in a given window size. We experimented AWE using fastText word embeddings with a window of size 5. We report results in Table~\ref{tbl:exp_wsd_verb}. BERT-based models set the new state of the art on this task, with the best results achieved by \cmbert and \frbertlarge.

\begin{table}[!htbp]
	\centering
	\begin{tabular}{lc}
    		\toprule
		Model & $F_1$  \\
		\midrule
		fastText & 34.90 \\
		mBERT & 49.83 \\
		\cmbert & 50.02 \\
		\frbertbase & 43.92 \\
		\frbertlarge & \textbf{50.48} \\
		\bottomrule
	\end{tabular}
	\caption{$F_1$ scores (\%) on the Verb Disambiguation Task.}
	\label{tbl:exp_wsd_verb}
\end{table}

\paragraph{Noun Sense Disambiguation}

We implemented a neural classifier similar to the classifier presented by \newcite{vialhal02131872}. This classifier forwards the output of a pre-trained language model to a stack of 6~trained Transformer encoder layers and predicts the synset of every input words through softmax. The only difference between our model and \newcite{vialhal02131872} is that we chose the same hyper-parameter as \frbertbase{} for the $d_{ff}$ and the number of attention heads of the Transformer layers (more precisely, $d_{ff}=3072$ and $A=12$).

\begin{table}[!htbp]
	\centering
	\begin{tabular}{lccc}
		\toprule
		Model & \multicolumn{2}{c}{Single} & Ensemble \\
		 & Mean & Std & \\
		\midrule
		No pre-training & 45.73 & $\pm$1.91 & 50.03 \\
		fastText & 44.90 & $\pm$1.24 & 49.41 \\
		mBERT & 53.03 & $\pm$1.22 & 56.47 \\
		\cmbert & 52.06 & $\pm$1.25 & 56.06 \\
		\frbertbase & 51.24 & $\pm$1.33 & 54.74 \\
		\frbertlarge & 53.53 & $\pm$1.36 & \textbf{57.85} \\ 
		\bottomrule
	\end{tabular}
	\caption{$F_1$ scores (\%) on the Noun Disambiguation Task.}
	\label{tbl:exp_wsd_nouns}
\end{table}

At prediction time, we take the synset ID which has the maximum value along the softmax layer (no filter on the lemma of the target is performed). We trained 8 models for every experiment, and we report the mean results, and the standard deviation of the individual models, and also the result of an ensemble of models, which averages the output of the softmax layer. Finally, we compared \frbert with \cmbert, mBERT, fastText and with no input embeddings. We report the results in Table~\ref{tbl:exp_wsd_nouns}.
On this task and with these settings, we first observe an advantage for mBERT over both \cmbert{} and \frbertbase{}. We think that it might be due to the fact that the training corpora we used are machine translated from English to French, so the multilingual nature of mBERT makes it probably more fitted for the task. Comparing \cmbert{} to \frbertbase{}, we see a small improvement in the former model, and we think that this might be due to the difference in the sizes of pre-training corpora. 
Finally, with our \frbertlarge model, we obtain the best scores on the task, achieving more than 1 point above mBERT.

\section{Conclusion}

We present and release \frbert, a pre-trained language model for French.
\frbert was trained on a multiple-source corpus and achieved state-of-the-art results on a number of French NLP tasks, surpassing multi-lingual/cross-lingual models. \frbert is competitive with \cmbert \cite{martin2019camembert} -- another pre-trained language model for French -- despite being trained on almost twice as fewer text data.
In order to make the pipeline entirely reproducible, we not only release preprocessing and training scripts, together with \frbert, but also  provide a general benchmark for evaluating French NLP systems (FLUE). \frbert is also now supported by 
Hugging Face's \texttt{transformers} library.\footnote{\url{https://huggingface.co/transformers/}}
\section{Acknowledgements}
This work benefited from the \textit{`Grand Challenge Jean Zay'} program and was also partially supported by MIAI@Grenoble-Alpes (ANR-19-P3IA-0003).

We thank Guillaume Lample and Alexis Conneau for their active technical support on using the XLM code.

\setcounter{section}{0}
\renewcommand{\thesection}{\Alph{section}}
\renewcommand{\thesubsection}{A.\arabic{subsection}}
\renewcommand{\thesubsection}{\thesection.\arabic{subsection}}
\section{Appendix}
\subsection{Details on our French text corpus}
Table~\ref{tbl:data_stats} presents the statistics of all sub-corpora in our training corpus. We give the description of each sub-corpus below.

\label{appx:desc_datasets}

\begin{table*}[htb!]
	\resizebox{\textwidth}{!}{
		\centering
		\begin{tabular}{lrrr}
			\centering
			\textbf{Dataset} & \textbf{Post-processed text size} & \textbf{Number of Tokens (Moses)} & \textbf{Number of Sentences} \\ [1ex] 
			\toprule
			
			CommonCrawl \cite{Buck-commoncrawl} & 43.4 GB & 7.85 B & 293.37 M \\ [1ex] 
			
			NewsCrawl \cite{li2019findings} & 9.2 GB & 1.69 B & 63.05 M \\ [1ex] 
			
			Wikipedia \cite{wiki:xxx} & 4.2 GB & 750.76 M & 31.00 M \\ [1ex] 
			
			Wikisource \cite{wiki:xxx}  & 2.4 GB & 458.85 M& 27.05 M\\ [1ex]
			
			EU Bookshop \cite{skadicnvsbillions} & 2.3 GB & 389.40 M & 13.18 M \\ [1ex] 
			
			MultiUN \cite{eisele2010multiun} & 2.3 GB & 384.42 M & 10.66 M \\ [1ex] 
			
			GIGA \cite{TIEDEMANN12.463} & 2.0 GB & 353.33 M & 10.65 M \\ [1ex] 
			
			PCT & 1.2 GB & 197.48 M & 7.13 M \\ [1ex] 
			
			Project Gutenberg & 1.1 GB & 219.73 M & 8.23 M \\ [1ex]
			
			OpenSubtitles \cite{lison2016opensubtitles2015}& 1.1 GB & 218.85 M & 13.98 M \\ [1ex] 
			
			Le Monde & 664 MB & 122.97 M & 4.79 M \\ [1ex] 
			
			DGT \cite{TIEDEMANN12.463} & 311 MB & 53.31 M & 1.73 M \\ [1ex] 
			
			EuroParl \cite{koehn2005europarl} & 292 MB & 50.44 M & 1.64 M \\ [1ex]
			
			EnronSent \cite{styler2011enronsent} & 73 MB & 13.72 M & 662.31 K \\ [1ex]
			
			NewsCommentary \cite{li2019findings} & 61 MB & 13.40 M & 341.29 K \\ [1ex] 
			
			Wiktionary \cite{wiki:xxx}  & 52 MB & 9.68 M & 474.08 K \\ [1ex]
			
			Global Voices \cite{TIEDEMANN12.463} & 44 MB & 7.88 M & 297.38 K \\ [1ex] 
			
			Wikinews \cite{wiki:xxx}  & 21 MB& 3.93 M & 174.88 K \\ [1ex]
			
			TED Talks \cite{TIEDEMANN12.463} & 15 MB & 2.92 M & 129.31 K \\ [1ex]
			
			Wikiversity \cite{wiki:xxx}  & 10 MB &  1.70 M & 64.60 K  \\ [1ex]
			
			Wikibooks \cite{wiki:xxx}  & 9 MB & 1.67 M & 65.19 K \\ [1ex]
			
			Wikiquote \cite{wiki:xxx}  & 5 MB & 866.22 K & 42.27 K \\ [1ex]
			
			Wikivoyage \cite{wiki:xxx}  & 3 MB & 500.64 K & 23.36 K \\ [1ex]
			
			EUconst \cite{TIEDEMANN12.463} & 889 KB & 148.47 K & 4.70 K \\ [1ex]
			\midrule
			\textbf{Total} & \textbf{71 GB} & \textbf{12.79 B} & \textbf{488.78 M} \\ [1ex]
			\bottomrule[0.3ex]
		\end{tabular}
	}
    \captionsetup{justification=centering}
	\caption{\label{tbl:data_stats}Statistics of sub-corpora after cleaning and pre-processing an initial corpus of 270 GB, ranked in the decreasing order of post-processed text size.}
\end{table*}

\paragraph{Datasets from WMT19 shared tasks}
We used four corpora provided in the WMT19 shared task \cite{li2019findings}.\footnote{\url{http://www.statmt.org/wmt19/translation-task.html}}
\begin{itemize}
	\item \textit{Common Crawl} includes text crawled from billions of pages in the internet. 
	\item \textit{News Crawl} contains crawled news collected from 2007 to 2018. 
	\item \textit{EuroParl} composes text extracted from the proceedings of the European Parliament.
	\item \textit{News Commentary} consists of text from news-commentary crawl. 
\end{itemize}

\paragraph{Datasets from OPUS} 
OPUS\footnote{\url{http://opus.nlpl.eu}} is a growing resource of freely accessible monolingual and parallel corpora \cite{TIEDEMANN12.463}. We collected the following French monolingual datasets from OPUS.

\begin{itemize}
	\item \textit{OpenSubtitles} comprises translated movies and TV subtitles.
	
	\item \textit{EU Bookshop} includes publications from the European institutions.
	
	\item \textit{MultiUN} composes documents from the United Nations.
	
	\item \textit{GIGA} consists of newswire text and is made available in WMT10 shared task.\footnote{\url{https://www.statmt.org/wmt10/}}
	
	\item \textit{DGT} contains translation memories provided by the Joint Research Center.
	
	\item \textit{Global Voices} encompasses news stories from the website Global Voices.
	
	\item \textit{TED Talks} includes subtitles from TED talks videos.\footnote{\url{https://www.ted.com}}
	
	\item \textit{Euconst} consists of text from the European constitution.
\end{itemize}

\paragraph{Wikimedia database} This includes Wikipedia, Wiktionary, Wikiversity, \etc The content is built collaboratively by volunteers around the world.\footnote{\url{https://dumps.wikimedia.org/other/cirrussearch/current/}}

\begin{itemize}
	\item \textit{Wikipedia} is a free online encyclopedia including high-quality text covering a wide range of topics. 
	\item \textit{Wikisource} includes source texts in the public domain.
	\item \textit{Wikinews} contains free-content news.
	\item \textit{Wiktionary} is an open-source dictionary of words, phrases \etc
	\item \textit{Wikiversity} composes learning resources and learning projects or research.
	\item \textit{Wikibooks} includes open-content books.
	\item \textit{Wikiquote} consists of sourced quotations from notable people and creative works.
	\item \textit{Wikivoyage} includes information about travelling.
\end{itemize}

\paragraph{Project Gutenberg} This popular dataset contains free ebooks of different genres which are mostly the world's older classic works of literature for which copyright has expired. 

\paragraph{EnronSent} This dataset is provided by \cite{styler2011enronsent} and is a part of the Enron Email Dataset,\footnote{\url{https://www.cs.cmu.edu/~enron/}} a massive dataset containing 500K messages from senior management executives at the Enron Corporation.

\paragraph{PCT} This sub-corpus contains patent documents collected and maintained internally by the GETALP\footnote{\url{http://lig-getalp.imag.fr/en/home/}} team.

\paragraph{Le Monde} This is also collected and maintained internally by the GETALP team, consisting of articles from Le Monde\footnote{\url{https://www.lemonde.fr}} collected from 1987 to 2003.

\section*{Bibliographical References}
\label{main:ref}
\bibliographystyle{lrec}
\bibliography{references}


\end{document}